\pdfoutput=1
\documentclass[letterpaper, 10 pt, conference]{ieeeconf}  

\IEEEoverridecommandlockouts                              

\usepackage{graphics} 
\DeclareGraphicsExtensions{.pdf,.png,.jpg}

\usepackage{epsfig} 
\usepackage{mathptmx} 
\usepackage{times} 
\usepackage{amsmath} 
\usepackage{amssymb}  
\usepackage{subfigure}
\usepackage[cjk]{kotex}
\usepackage{cite}
\usepackage{mwe}
\usepackage{xcolor}
\usepackage{url}
\usepackage{multirow}
\usepackage{fixltx2e}
\usepackage{tabularx}

\def\BigRoman{\uppercase\expandafter{\romannumeral\number\count 255 }}
\def\Romannumeral{\afterassignment\BigRoman\count255=}

\title{\LARGE \bf
BRM Localization: UAV Localization in GNSS-Denied Environments Based on Matching of Numerical Map and UAV Images
}

\author{Junho Choi$^{1}$ and Hyun Myung$^{2}$, \textit{Senior Member, IEEE}
\thanks{*This work was supported by BK21+ and a grant (contract num.: 20200302731-00) from the Drone Safety and Activation Support Consignment Project of the Korea Institute of Aviation Safety Technology (KIAST).}
\thanks{$^{1}$Junho Choi and $^{1}$Hyun Myung are with the Urban Robotics Laboratory, Korea Advanced Institute of Science and Technology(KAIST) Daejeon, 34141, South Korea. Prof. Hyun Myung is with School of Electrical Engineering, KI-AI, and KI-R at KAIST.
        {\tt\small cjh6685kr@kaist.ac.kr, hmyung@kaist.ac.kr}}}%

\begin{document}

\maketitle
\thispagestyle{empty}
\pagestyle{empty}

\begin{abstract}

Localization is one of the most important technologies needed to use Unmanned Aerial Vehicles (UAVs) in actual fields. Currently, most UAVs use GNSS to estimate their position. Recently, there have been attacks that target the weaknesses of UAVs that use GNSS, such as interrupting GNSS signal to crash the UAVs or sending fake GNSS signals to hijack the UAVs. To avoid this kind of situation, this paper proposes an algorithm that deals with the localization problem of the UAV in GNSS-denied environments. We propose a localization method, named as BRM (Building Ratio Map based) localization, for a UAV by matching an existing numerical map with UAV images. The building area is extracted from the UAV images. The ratio of buildings that occupy in the corresponding image frame is calculated and matched with the building information on the numerical map. The position estimation is started in the range of several \textit{km$^{2}$} area, so that the position estimation can be performed without knowing the exact initial coordinate. Only freely available maps are used for training data set and matching the ground truth. Finally, we get real UAV images, IMU data, and GNSS data from UAV flight to show that the proposed method can achieve better performance than the conventional methods.

\end{abstract}

\section{INTRODUCTION}

Unlike ground robots, aerial vehicles have an advantage of being able to move freely in three-dimensional environments, so it is emerging as an alternative to overcome the limitations of ground robots. As the technological level of sensor system, flight control, and communication system has advanced, the use of UAVs has been expanded in the last few years, and many attempts have been made to apply autonomous UAVs to actual fields\cite{UAV_application1,UAV_application2,UAV_application3,UAV_application4,UAV_application5, UAV_application6}. It is essential to accurately estimate the position of the UAV in autonomous flight because the autonomous flying UAV determines a flight path based on the current position and performs a given task. In general, UAVs flying outdoors recognize their location based on GNSS. This method has the advantage that the absolute coordinates can be estimated with error from several meters to several centimeters, depending on the GNSS receiver device used. However, if the UAV localizes based only on GNSS, it has a critical limitation that it can be used only in an environment where signals are well received. GNSS signals are easily blocked by buildings and are vulnerable to disturbances. Additionally, UAVs using GNSS signals can be attacked by jamming and spoofing. Therefore, UAV localization that relies only on GNSS signals is not a robust method. In this paper, we propose an image-based localization method to estimate the global position of the UAV without using GNSS.

\subsection{Related Works}
The research for UAV localization without GNSS signals is actively being conducted. Various types of sensors have been tried to accurately estimate the location of drones in large areas. A UAV has a limited payload due to the characteristics of its platform, and if the payload is high or a sensor that consumes a lot of power is used, the flight time is shortened. For this reason, lightweight cameras with low power consumption are widely used. VIO (Visual Inertial Odometry) algorithm has been studied to reduce the error when using only a camera with an IMU\cite{delmerico2018benchmark,vins-mono,svo1,svo2,rovio,okvis}. Since this method estimates the relative position of a UAV after its initialization, there is a limitation that the absolute position cannot be known. If the UAV flies in a path without loop closing, errors will accumulate. Also, if the UAV loses its position, it is difficult to recognize the relative position from the initial position. Various methods have been carried out to solve this problem. A method using a marker such as a QR code\cite{qrcode} or an April Tag\cite{apriltag} has been proposed\cite{marker}. It is a method of calibrating the position of a UAV when a corresponding marker is recognized after pre-positioning a specific marker. This method has the advantage of being able to estimate the position very accurately once the marker is recognized. However, this localization method is not suitable for UAVs flying over large areas. There has been an attempt to develop absolute position estimation method based on the local feature on the flight path\cite{local_feature}, a method that learns landmarks such as buildings located near the flight path in advance and estimates the position of the UAV based on the learned landmark. Nevertheless, this method also has a problem that localization becomes difficult when the UAV deviates from the learned path.

Recently, methods that use free available maps have been proposed to solve this problem\cite{nassar2018deep,goforth2019gps,yol2014vision}. In particular, satellite maps, which store very large areas in the form of orthogonal images, are frequently used to estimate the location of UAVs. Many methods for localization of the UAVs by matching satellite maps with images obtained from the camera which is attached downward from the UAV have been studied\cite{goforth2019gps, nassar2018deep, yol2014vision}. In the case of satellite maps, the image was taken at a certain point of time at a height of hundreds to thousands of kilometers. To match the images taken from the UAV to satellite images, it is necessary to resolve perspective effects, lighting condition differences, seasonal differences, and regional variations. Feature-based matching methods such as SIFT\cite{SIFT}, SURF\cite{surf}, ORB\cite{ORB}, etc. fail to match as shown in Fig. 1.
 
   \begin{figure}[h]
      \centering
      \includegraphics[scale=0.4]{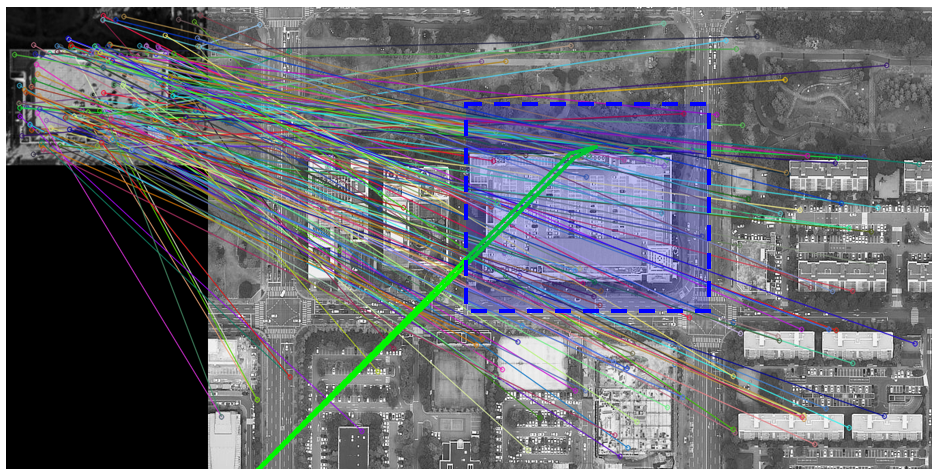}
      \caption{This figure shows that SIFT algorithm, which is the widely used conventional point feature matching, fails in matching satellite images(right) and UAV images(left). The area marked in blue is the same area as the image on the left, but the SIFT algorithm fails to find the blue area.}
      \label{figurelabel}
   \end{figure}

In \cite{goforth2019gps}, they used the idea that matching satellite images taken at different times would be a similar task to matching satellite and UAV images. They proposed a localization algorithm by matching satellite maps and UAV images after learning several identical satellite maps taken at various times to consider seasonal, weather, and light disparities. This algorithm works at various environments, from gravel pits to villages. However, the limitation is that the initial position of the UAV should be exactly given and that the algorithm was only tested in a narrow area. In addition, the algorithm was verified only in a very stable situation by capturing the image along a straight path from the static 3D reconstructed map rather than from the actual flight. In \cite{nassar2018deep}, they proposed an SSM (Semantic Shape Matching) method. Some classes, which have specific shapes such as buildings and roads, are segmented from both satellite maps and UAV images and matched by comparing contours of each segmented areas. However, it is also assumed that the GPS position of the initial starting position is known. Additionally, since the shape of the segmented area is used for matching, the algorithm is highly dependent on segmentation accuracy, which makes it difficult to relocalize if the building is lost in complex areas such as the city center. In \cite{yol2014vision}, the measurement of statistical dependency between two signals (mutual information) has been proven to work more robustly than the sum of squared differences (SSD) which is widely used in vision-based matching algorithms. However, because the experiment was only done in a region as small as 130m $\times$ 100m, it is unclear whether the algorithm works well when flying over long distances or when the environment changes.

\subsection{Contributions}
In this paper, we propose an algorithm for estimating the absolute position of the UAV by matching UAV images and a numerical map without using GNSS. Unlike the previous papers, we propose a localization algorithm for UAVs in the absence of exact position coordinates at the starting point of flight. However, it is assumed that the UAV performs the mission within the designated area and exists within the map of the area. For localization, we propose the building ratio feature. Building area is segmented by a segmentation neural network. These segmented images of the buildings are divided into circular areas with various radii, and the ratio of the building area to the circular area is used as a feature. This proposed feature is rotation invariant so it has the advantage of estimating the position candidate group without knowing the position and the orientation. Additionally, it is possible to start localization again at any time even if the position is lost. This feature also works more robustly in terms of the segmentation accuracy than other shape matching algorithms. The training data sets and maps for matching are just gained from freely available online data. The previous satellite map matching-based algorithms are tested only at static environments or in narrow areas. However, we verify the feasibility of the proposed algorithm by real flight data acquired from a UAV.

\subsection{Overview}
In Section {\Romannumeral 2}, a framework of the BRM (Building Ratio Map based) localization is introduced. Then, the proposed feature is defined. Also, a process of the building ratio map generation is described. Building ratio matching algorithm is explained in Section {\Romannumeral 3}. We explain how the number of candidates decreases and that the global position is estimated. In Section {\Romannumeral 4}, we analyze the results of the proposed algorithm and the VIO algorithm. Finally, we conclude with the future works in Section {\Romannumeral 5}.


\section{Proposed Feature and Building Ratio Map}
\subsection{Proposed Algorithm Framework}
   \begin{figure}[h]
      \centering
      \includegraphics[scale=0.41]{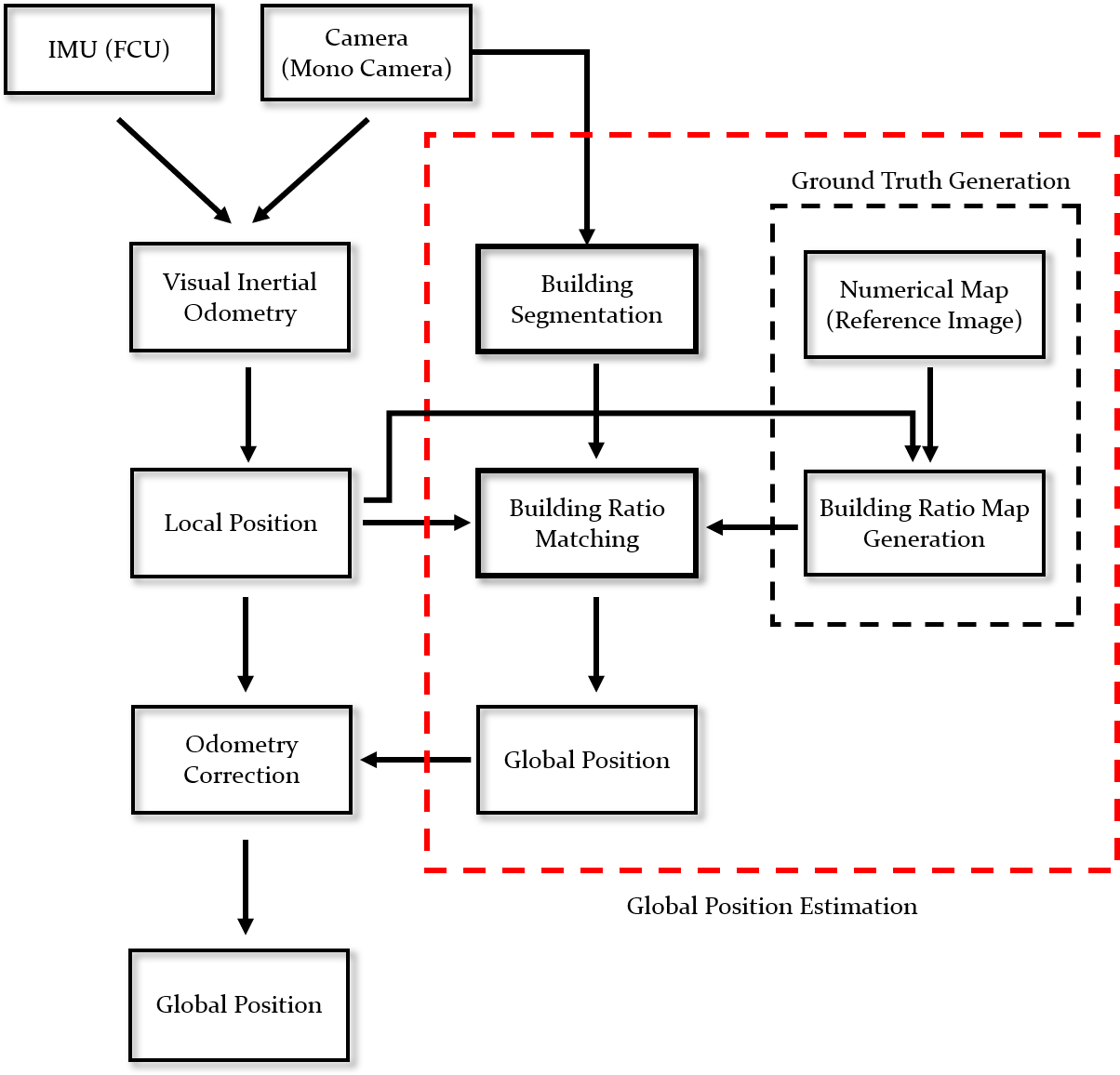}
      \caption{Flow of the proposed localization algorithm. IMU and image data are received from a flight controller unit and a camera. From these two inputs, a local odometry is calculated. UAV images are converted to segmented building images. Then, the building ratio matching algorithm runs and the global position of the UAV is estimated.}
      \label{figurelabel}
   \end{figure}
The whole algorithm is shown in Fig. 2. IMU data is obtained from the Flight Control Unit (FCU) and RGB images are received from the monocular camera mounted downward on the UAV. The Visual Inertial Odometry algorithm is operated on two inputs to estimate the relative position from the initial position of the UAV and the relative odometry between each image frame. Images from the camera are passed through the segmentation network at regular intervals and they are converted to segmented building images. The features for matching are calculated from the segmented building images. Each image has not only feature values but also odometry constraints on the previous images. In the pre-processed numerical map, location candidates that meet these conditions are deduced. After the convergence of candidates, the global position of the UAV is estimated.

\subsection{Building Segmented Images}
$w \times h$ $(w > h)$ sized RGB images are received from the mono camera. From this image sequence, some images are extracted at regular intervals to be passed through the pre-trained segmentation network. Through the network, the building area is segmented. To create a rotation invariant feature from the original image, only the $h \times h$ square area that is obtained by cutting off both ends of the original image is used. The \textit{i}-th image created in this way is denoted as $I_{i}$ $(i=1, 2,\dots,m)$.

\subsection{Building Ratio Feature}
   \begin{figure}[h]
      \centering
      \includegraphics[scale=0.5]{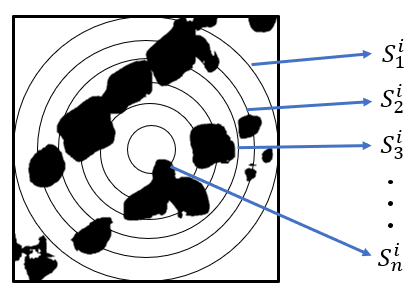}
      \caption{The proposed building ratio feature. Black area is the building segmented area. The ratios of building area in the certain circular areas are used as features. Each circular area is the area covered by all the region until the center.}
      \label{figurelabel}
   \end{figure}

   For the situation where there is no initial information about the location, the following features are proposed for the localization as shown in Fig. 3. The image $I_i$ is divided into \textit{n} circular regions which are denoted by \textit{$S_{k}^{i}$} \textit{$(k=1,\dots,n)$}. \textit{$S_{k}^{i}$} is an area with the radius \textit{$r_{k}$} sharing a center point with $I_i$.
   \begin{equation}
r_{k} = \cfrac{h}{2}\cdot\cfrac{n+1-k}{n}
\end{equation}
   The area of the buildings (black area in Fig. 3) in the area \textit{$S_{k}^{i}$} is defined as \textit{$B_{k}^{i}$}, where the ratio of \textit{$B_{k}^{i}$} to \textit{$S_{k}^{i}$} is used as a feature as follows:
   
\begin{equation}
f_{k}^{i} = \cfrac{B_{k}^{i}}{S_{k}^{i}}.
\end{equation}
Then, each $I_i$ has \textit{n} features. Since \textit{$f_{k}^{i}$} is a rotation invariant feature, it is possible to estimate the position of the UAV without the initial position based only on the RGB image taken from the UAV and the local odometry estimated based on visual inertial odometry algorithm.

\subsection{Building Ratio Map Generation}
\begin{figure}
\centering  
\subfigure[]{\includegraphics[width=0.48\linewidth]{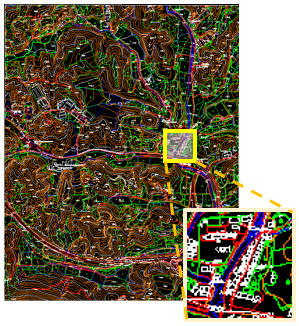}}
\subfigure[]{\includegraphics[width=0.48\linewidth]{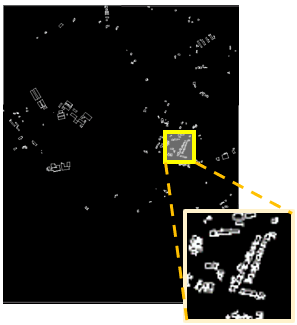}}
\subfigure[]{\includegraphics[width=0.48\linewidth]{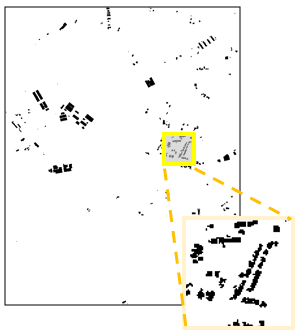}}
\subfigure[]{\includegraphics[width=0.48\linewidth]{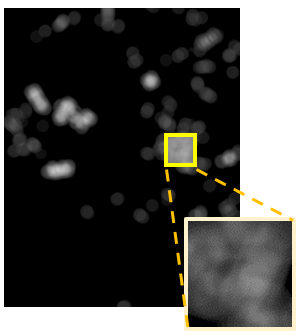}}
\caption{The process of building ratio map generation. (a) original numerical map. (b) numerical map which contains only building related layer. (c) binary numerical map where the building area is black part and the other part is white. (d) building ratio map which is the reference feature map.}
\end{figure}

Previously, it is described that the building ratio at predetermined areas of $I_i$ is used as a feature. In addition, we already know the building information of each location through the numerical map. Therefore, when the UAV is located at a certain position within the numerical map, it is possible to know in advance what feature values are to be calculated in the image $I_i$.
In other words, the true value of \textit{$f_{k}^{i}$} is known. We define a building ratio map \textit{$M_{k}$} which consists of the true value of \textit{$f_{k}^{i}$} which is denoted by \textit{$\tilde{f}_{k}$}.
 The example of the original numerical map is shown in Fig. 4 (a). This map includes various artifacts and natural topography. We use building information from this map as shown in Fig. 4 (b), which is composed of the exact position and outline information of buildings. Therefore, this map is used for generating the ground truth. Fig. 4 (c) is a binary map which is generated from Fig. 4 (b). It is a numerical map divided into a building area (black) and a non-building area (white). Finally, the building ratio map \textit{${M}_{k}$} is generated as shown in Fig. 4 (d).
 This building ratio map is generated as follows. First of all, the window size on the numerical map $\tilde{S}_{1}^{i}$ is calculated to be the same scale with $I_i$ as follows:
\begin{equation}
\tilde{S}_{k}^{i} = \pi(\cfrac{n+1-k}{n}\textit{$z_{l}$}\tan{\cfrac{\alpha}{2}})^{2}
\end{equation}
 where \textit{$z_{l}$} is the height of the UAV calculated from visual inertial odometry and $\alpha$ is field of view of the camera. The same number of maps as the number of features used are generated.


\section{Building Ratio Matching Algorithm}
\subsection{Preliminaries}
\begin{figure}[h]
\centering  
\subfigure[]{\includegraphics[width=0.32\linewidth]{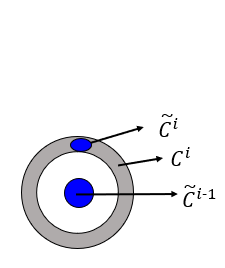}}
\subfigure[]{\includegraphics[width=0.32\linewidth]{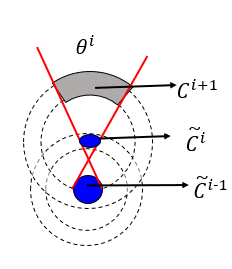}}
\subfigure[]{\includegraphics[width=0.32\linewidth]{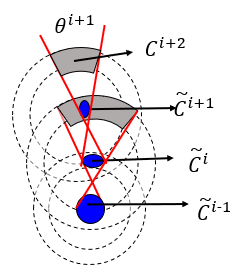}}
\caption{Examples of candidates. The gray areas are candidates based on the movement of the UAV. The blue areas are candidates which are calculated by matching with the building ratio map. (a) example of candidates when $\tilde{C}^{i-2}$ = $\emptyset$. (b) $C^{i+1}$ is selected by the constraints of the UAV movement. (c) $\tilde{C}^{i+1}$ is calculated by matching with the building ratio map and $C^{i+2}$ is selected by the constraints of the UAV movement.}
\end{figure}

In the following, we define some notations which will be used in the next section. First of all, ${C}^{i}$ is a set of \textit{i}-th candidates that are selected by the constraints of the UAV movement such as the travelled distance and heading angle. These candidates are depicted in gray color in Fig. 5. $\tilde{C}^{i}$ is a set of \textit{i}-th candidates that are calculated by matching the building ratio map with the feature $f_{k}^{i}$ $(k=1,2,\dots,n)$. These candidates are represented in blue color in Fig. 5. Local odometry of the UAV is calculated during the flight. So two constraints of the UAV movement between two positions, heading and traveled distance of the UAV are made. $\theta^{i}$ is the heading constraint of the UAV at $I_{i}$. It is described in Fig. 5 (b) and (c) and calculated by (4) where $\theta$ is as follows:
\begin{equation}
\theta^{i-1} = \left\{\theta \mid \theta = \arctan({\cfrac{y_{i-1}-y_{i-2}}{x_{i-1}-x_{i-2}}})\right\}
\end{equation}
where $(x_{i-1},y_{i-1})$ and $(x_{i-2},y_{i-2})$ are arbitrary points of the (\textit{i}-1)-th candidates  $\tilde{C}^{i-1}$  and (\textit{i}-2)-th candidates $\tilde{C}^{i-2}$, respectively. If $\tilde{C}^{i-2} = \emptyset$, then $\theta$ can be any degrees between 0 to 360.

\subsection{Candidate Selection}
  \begin{figure}[h]
      \centering
      \includegraphics[scale=0.35]{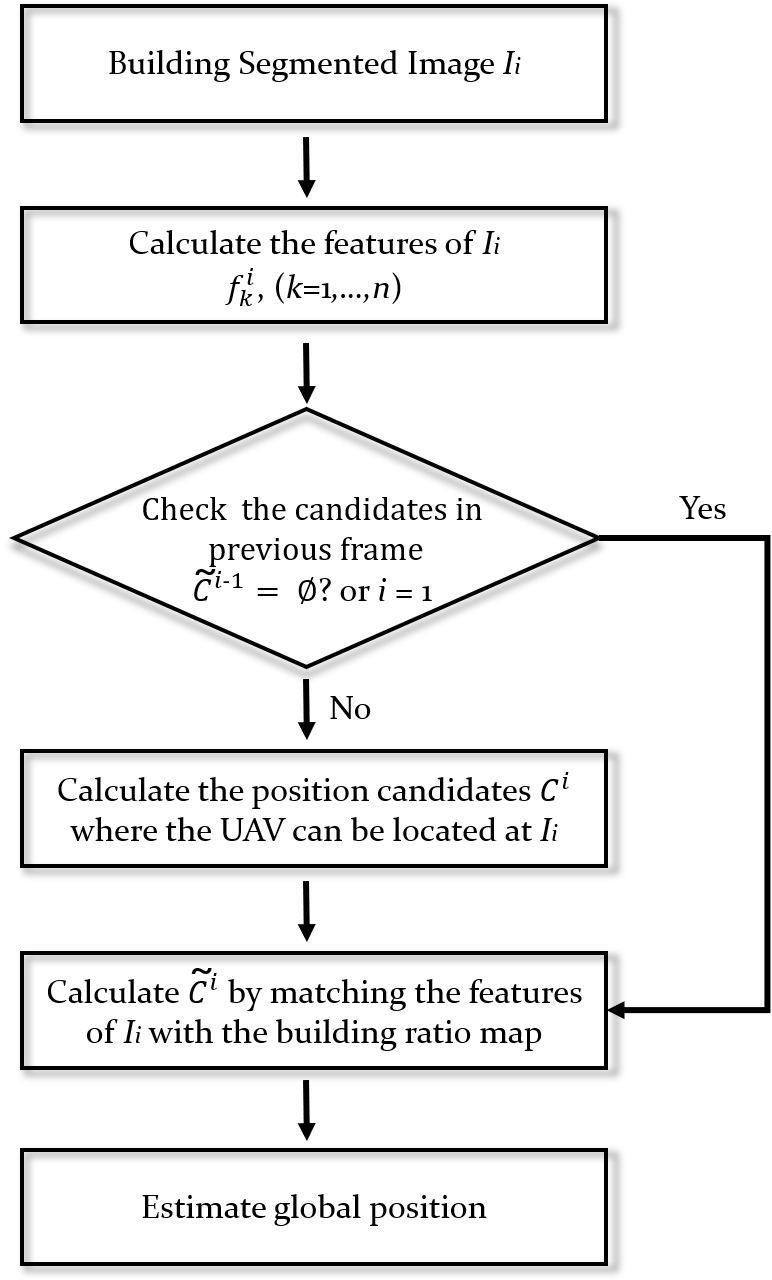}
      \caption{
Flow chart of the building ratio matching algorithm. When \textit{i} is received, the feature \textit{$f_{k}^{i}$} is calculated. Then, the previous candidates $\tilde{C}^{i-1}$ are checked for absence. If there is no candidate, new candidate $C^{i}$ is searched throughout the whole building ratio map. If candidates exist, the candidates $C^{i}$ in the current image is determined. After that, the candidates $\tilde{C}^{i}$ for the current location is recalculated by matching with the building ratio map. When the candidates are calculated, the global position of the UAV is estimated.
}
      \label{figurelabel}
   \end{figure}
   
The flow chart of the proposed building ratio matching algorithm is described in Fig. 6. When the segmented building image $I_i$ is generated by the method described in Section {\Romannumeral 2.\textit{B}}, feature \textit{$f_{k}^{i}$} is calculated. Before estimating the candidates of the current position, the candidates at $I_{i-1}$, which are denoted by $\tilde{C}^{i-1}$, are checked. This is because the previous candidates impose constraints on the areas where the current candidates may exist. According to the presence or absence of the position candidates in the preceding image frame, the position estimation algorithm is divided into two. 
If $I_{i}$ is the first image or all the previous candidates were eliminated, there are no candidates of the previous position. In this case, we can assume that all positions on the map can be the initial candidates of $I_{i}$ because $C^{i}$ has no movement constraint. Then
\begin{equation}
C^{i} = \{(x,y) \mid (x,y) \in (X_{M},Y_{M})\}
\end{equation}
where ($X_{M}$, $Y_{M}$) is a set of coordinates on the numerical map. And candidates are calculated on the entire map as follows:
\begin{equation}
\tilde{C}^{i} = \{(x,y) \in C^{i} \mid \sum_{k=1}^{n}|M_{k}(x,y)-f_{k}^{i}| < e_{1}\}
\end{equation}
where $M_{k}(x,y)$ is the true feature value $\tilde{f}_{k}$ at $(x,y)$ position and $e_{1}$ is the feature error threshold. 
If the previous candidates exist, $C^{i}$ is estimated from the distance traveled and the heading angle of the UAV as follows:
\begin{equation}
\begin{aligned}
C^{i} & = \{(x,y) \mid |x-(x_{i-1}+d^{i}\cdot\cos{\theta^{i-1}})| < \epsilon \\
& \qquad \qquad,  |y-(y_{i-1}+d^{i}\cdot\sin{\theta^{i-1}})| < \epsilon\}
\end{aligned}
\end{equation}
where $\epsilon$ is the distance error constant and $d^{i}$ is the distance between $I_{i-1}$ and $I_{i}$ which is estimated by VIO. After the constraints-based candidates ${C}^{i}$ are selected, $\tilde{C}^{i}$ are calculated in the same way as (6). Whenever the maximum distance between the average point of $\tilde{C}^{i}$ and an arbitrary point of $\tilde{C}^{i}$ is less than the constant $d_{max}$, we can determine the convergence and the global position can be estimated by averaging $\tilde{C}^{i}$.


\section{Result}
\subsection{Implementation Details}
\textbf{Hardware configuration}
  \begin{figure}[h]
      \centering
      \includegraphics[scale=0.25]{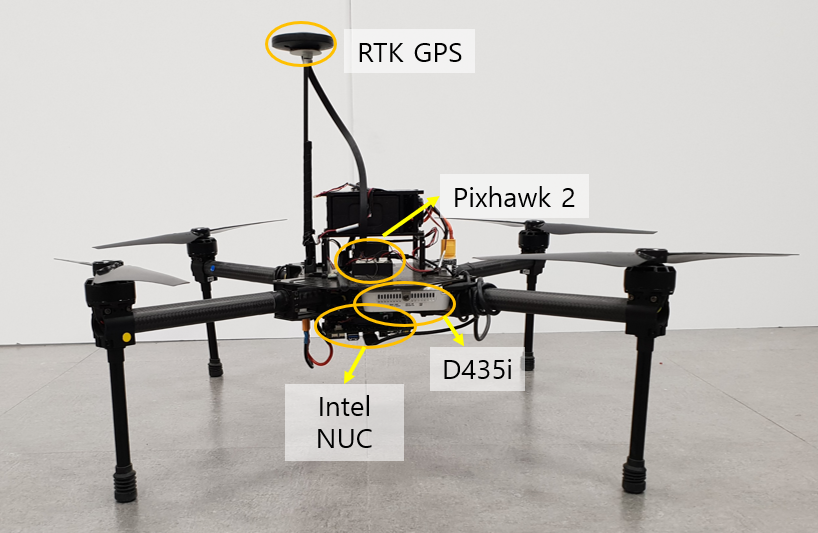}
      \caption{The UAV configuration. DJI Matrice 100 is used as the UAV frame, and Pixhawk2 is used for flight control and sensor information collection. Downward images are acquired with the D435i mono-camera. Intel NUC is used for VIO algorithm testing, images storage, and other calculations. RTK GPS is used to obtain ground truth of the UAV flight path.}
      \label{figurelabel}
   \end{figure}

The UAV configuration is shown in Fig. 7. We used DJI Matrice 100 for the drone platform. In selecting the UAV, we considered whether it can hold the wind enough, sustain enough flight time, and be loaded with the necessary sensors on top of it while being not too large or heavy. The monocular camera of Realsense D435i is used. The camera was chosen because it is light enough (72 grams) and the RGB mono camera uses a global shutter. In the case of the camera using the rolling shutter, the screen distortion occurs when the vibration occurs in the actual flight, and the visual inertial odometry algorithm does not work well. We use the Pixhawk2 for flight controller which is capable of obtaining 100Hz IMU sensor data and other sensor information. Intel NUC is used to verify the VIO algorithm result during flight and to save bag files (experimental data storage). The actual flight route is obtained via Here+ RTK GPS. Since RTK GPS has a position accuracy in the range of several cm, the actual flight path is recorded and used to obtain data for comparison with the path estimated by the proposed algorithm.

\textbf{Training network}
In the EO (Earth Observation) task that extracts the information from the satellite map, such as building segmentation and road segmentation, it was observed that U-Net\cite{u-net} showed the best performance\cite{u-net_comparison, u-net_comparison2}, thus U-Net is used for learning. The numerical map mentioned in Section {\Romannumeral 2} is processed and used as labeled data. The size of numerical map is $4,403\times5,555$ pixels with 0.5 meter-per-pixel resolution. The resolution of satellite images is $2,048\times2,048$ pixels. Scale of the satellite images are between 0.23 and 0.47 meter-per-pixel. All the data sets are downloaded from the cities that are not the actual experimental areas and we selected areas that include buildings, fields, rivers, and roads. Numerical maps and satellite maps were resized and cropped to $512\times512$ size and used for training. We trained with a learning rate of $7 \cdot 10^{-7}$ and 124 epoch. The train accuracy is 0.940 and the test accuracy is 0.916.

\textbf{Numerical map and satellite map}
Only freely available maps are used for localization and training data set generation. Satellite images are downloaded from Kakao Map\cite{kakao}. We use numerical map which is freely available at National Geographic Information Institute of Korea\cite{numerical}. The layers from B0014111 to B0014118 at the dxf file of the numerical map are used.

\subsection{Acquisition of the Test Data}
  \begin{figure}[h]
\centering  
\subfigure[]{\includegraphics[width=0.6\linewidth]{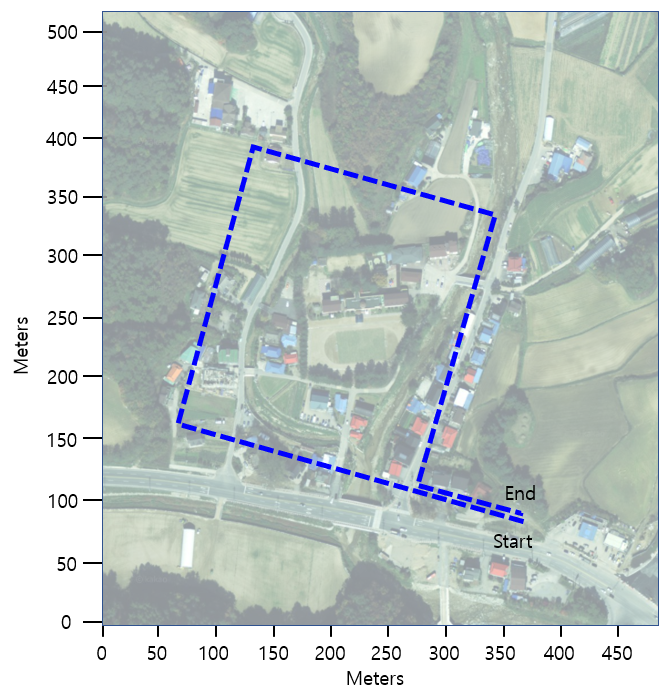}}
\subfigure[]{\includegraphics[width=0.13\linewidth]{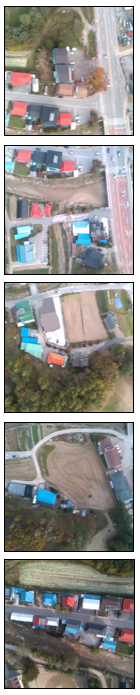}}
\subfigure[]{\includegraphics[width=0.13\linewidth]{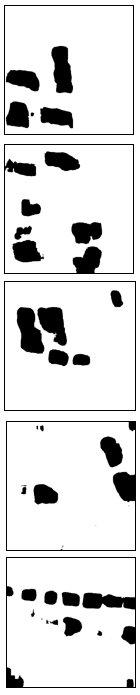}}
\caption{(a) The flight path. UAV took off at the start point and landed at the end point after flying along the square path. (b) Some examples of the image frames taken by the RGB camera during the UAV flight. (c) The images corresponding to (b) in which the buildings are segmented through the segmentation network.}
\end{figure}

\begin{table}[b]
\caption{Parameters used in the experiment}
\label{parameters}
\begin{center}
\begin{tabular}{|c|c|c|}
\hline
Parameter & Value & Meaning\\
\hline\hline
\textit{n} & 3 & 3 features are used in the experiment\\
\hline
$\alpha$ & $43{^\circ}$ & Field of view of the monocular camera\\
\hline
$z_{l}$ & 150m & the flight height is fixed to \textit{$z_{l}$}\\
\hline
$e_{1}$ & 0.3 & feature error threshold in (5)\\
\hline
$\epsilon$ & 25m & distance error constant in (6)\\
\hline
$d_{max}$ & 75m & convergence determination constant\\
\hline
\end{tabular}
\end{center}
\end{table}

  The experimental area is near Daegwallyeong, Pyeongchang, Gangwon-do, South Korea. Fig. 8 (a) shows the satellite images of the area where experiment was conducted. The blue line in Fig.8 (a) is the flight path. The flight was performed in a rectangular path and, to simulate the long distance flight, the experimental data was used under the condition in which loop closing algorithm in the VIO is not used. Fig.8 (b) shows some examples of the UAV images and Fig.8 (c) shows the corresponding building segmented images. The UAV flied at an average speed of 5m/s at a height of 150m. The total flight distance is 1,075m. The UAV images were acquired with 30Hz. From this image sequence, the building ratio matching algorithm was run every 5 sec. This is because the aircraft flied at an altitude of 150m and the aircraft had to move about 25m to make a meaningful change in the input images. Other parameters used are shown in Table {\Romannumeral 1}.

\begin{figure*}[h]
\centering  
\subfigure[]{\includegraphics[width=0.31\linewidth]{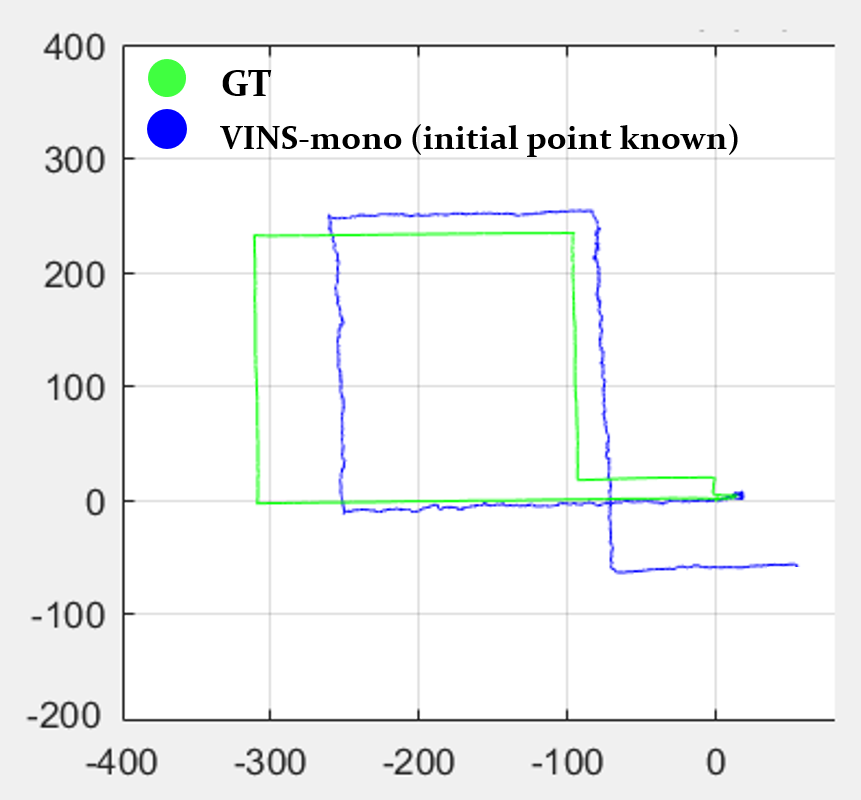}}
\subfigure[]{\includegraphics[width=0.31\linewidth]{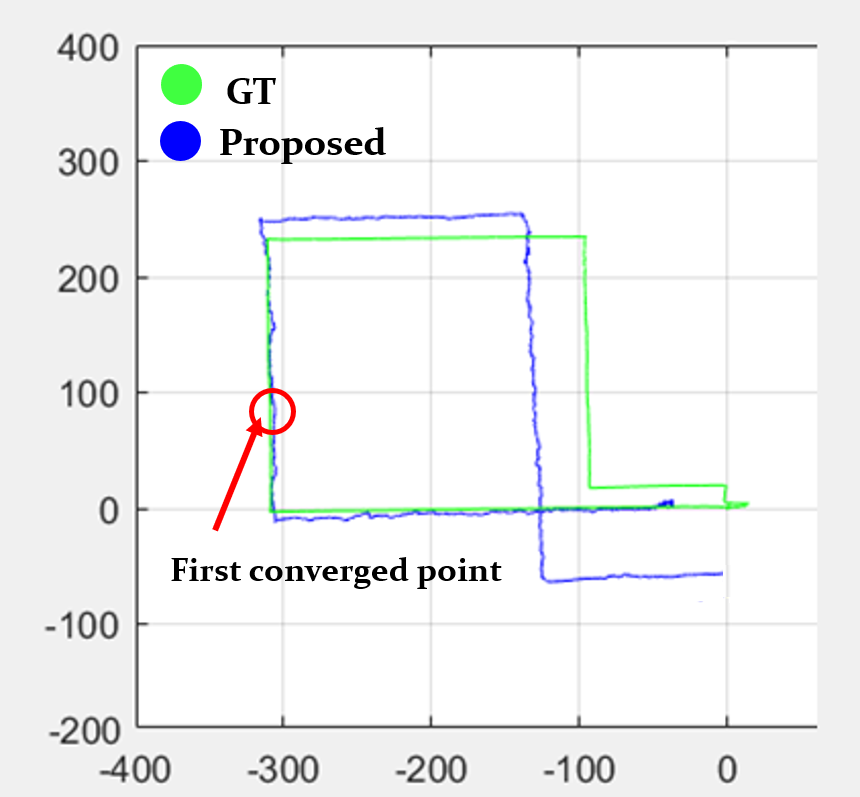}}
\subfigure[]{\includegraphics[width=0.31\linewidth]{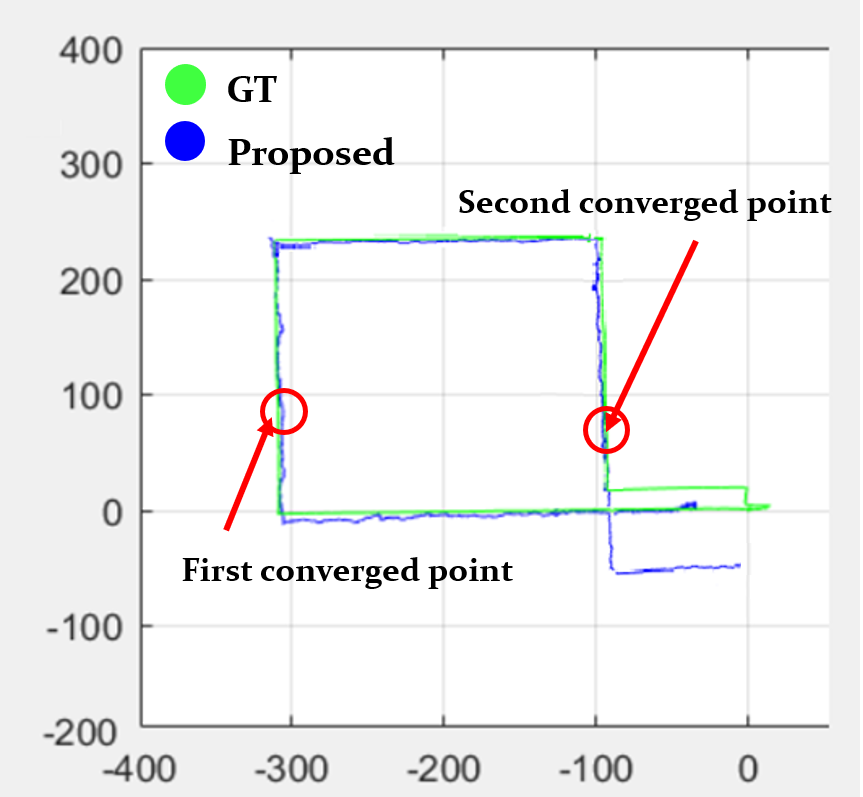}}
\caption{Results of the localization. Green line is the ground truth (trajectory of RTK GPS) and blue line is the estimated trajectory. (a) The trajectory estimated by VINS-mono algorithm. (b) The trajectory estimated by building ratio matching and VINS-mono. The UAV is localized at the red circle (First converged point). To compare with (c), only VINS-mono is used after the first converged point. (c) The trajectory estimated by building ratio matching and VINS-mono. The UAV is localized twice (two red circles).}
\end{figure*}

   \begin{figure}[h]
      \centering
      \includegraphics[scale=0.5]{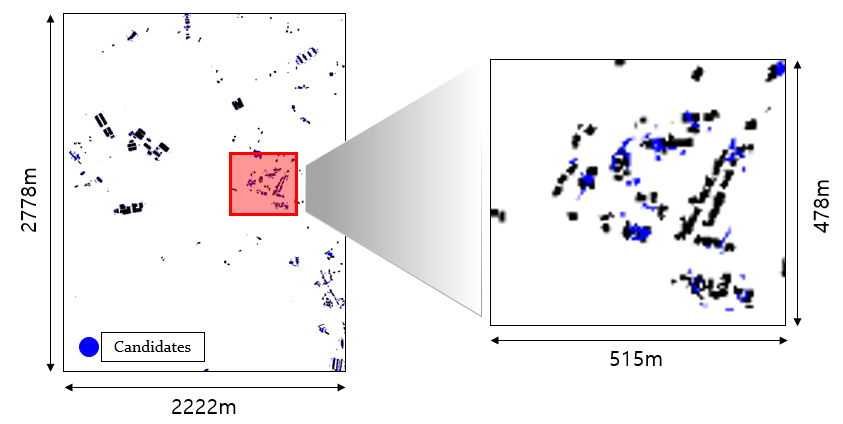}
      \caption{Result of the first candidates calculation (blue). The image on the left shows the candidates estimated at the entire map. The image on the right is an enlarged image of the actual flight area on the entire map.}
      \label{figurelabel}
   \end{figure}

\subsection{Experimental Results}

\begin{table}[b]
\caption{Root-Mean-Square-Error of each result}
\begin{center}
\begin{tabular}{|c c c c|}
\hline
RMSE & VINS-mono & Proposed 1 & Proposed 2 \\
\hline\hline
Whole path & 48.27m & 31.25m & \textbf{7.53m} \\
\hline
After first convergence & 79.58m & 77.12m & \textbf{12.01m} \\
\hline
\end{tabular}
\end{center}
\end{table}

Results of the VIO algorithm and proposed algorithm are shown in Fig. 9. The ground truth trajectory which is obtained from RTK GPS is represented in green color. The trajectories which are estimated by VIO and the proposed algorithm is represented in blue color. VINS-mono\cite{vins-mono} is used as a comparison algorithm. VINS-mono algorithm can not estimate the initial point, thus the start point was given to VINS-mono. Fig. 9 (a) is the trajectory estimated by only VINS-mono. A trajectory which uses only VINS-mono after the first convergence is generated as shown in Fig. 9 (b) for the fair comparison when both VINS-mono and the proposed algorithm when the initial point is known (TABLE {\Romannumeral 2} Proposed 1). Fig. 9 (c) is the trajectory of the proposed algorithm (TABLE {\Romannumeral 2} Proposed 2). In this case, the UAV calculated the candidates until it landed.

The RMSE values are shown in TABLE {\Romannumeral 2}. The whole length of the path is 1,075m. The RMSE of the whole path of Proposed2 which uses the proposed algorithm until it landed is significantly lower than VINS-mono result which are 48.27m and 7.53m. In the case of the initial points of both VINS-mono and the proposed algorithm are known, the RMSE of VINS-mono was 77.12m while RMSE of our proposed method was 12.01m.

      \begin{figure}[t]
      \centering
      \includegraphics[scale=0.55]{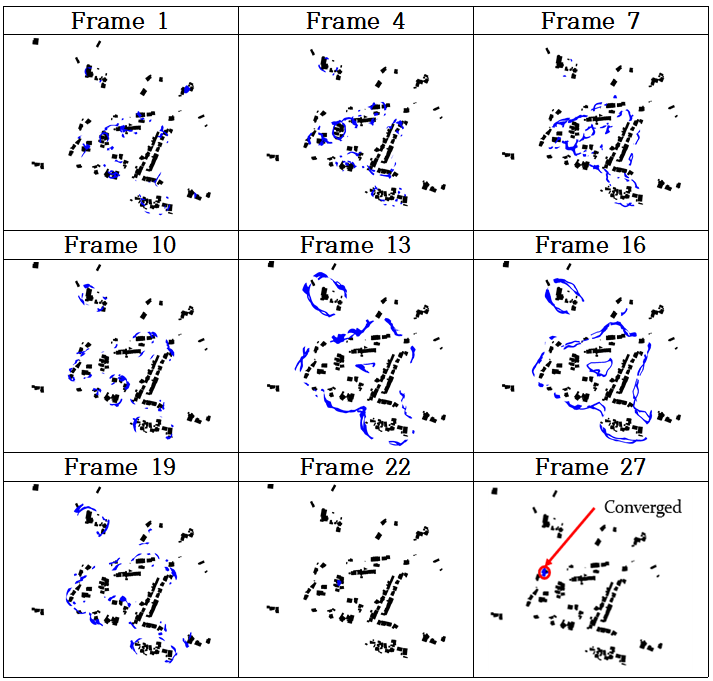}
      \caption{Some examples of the candidates calculation. This figure starts with the candidate estimated by the first image frame and shows how the candidates are converged.}
      \label{figurelabel}
   \end{figure}

   Fig. 10 shows the first candidates (blue) of the proposed algorithm. The image on the left is the entire numerical map with the candidates. The image on the right is an enlarged image of the actual flight area as shown in Fig. 8 (a). Some examples of the candidates at the actual flight area are described on Fig. 11. The candidates are initially scattered and converge at frame 27. The global position of the UAV is firstly estimated at this frame.

\section{CONCLUSIONS}
In this paper, we proposed a novel localization algorithm, named as BRM (Building Ratio Map based) localization, for a UAV based on building ratio matching using numerical map and the UAV images in GNSS denied environments. The conventional approaches need the initial position to estimate the absolute position of the UAV. However, our algorithm estimates the position of the UAV when the initial position is given within wide area e.g. \textit{$2.2km\times2.6km$}. Since the UAV is used to accomplish a given goal in a particular region, the assumption that the initial location is known in the range of \textit{$km^{2}$} is a resonable assumption and it can be extended to wider areas in the proposed algorithm. The proposed algorithm is verified using real UAV flight data which include UAV images, IMU data, and GPS data.

The experiment has been carried out in Pyeongchang, Gangwon-do. The error is analyzed by comparing with the actual flight path measured by RTK GPS. The candidates have converged twice during the flight with 6.1m and 7.8m errors, respectively, which are small enough since these are after 382m and 891m long flights at a height of 150m. It is confirmed that the RMSE (Root Mean Square Error) of the proposed algorithm is smaller than that of VINS-mono. In the absence of the initial position, the position is better estimated than VINS-mono, and after two convergences, the error about the entire path is much smaller than that of VINS-mono. For the fair comparison when the initial position is known, we analyze the position error after the convergence of candidates. This means that the errors of the proposed algorithm and VINS-mono are compared when both are provided with the starting position. In this case, the result shows that the RMSE is significantly reduced from that of VINS-mono.

In this paper, only building ratio information is used as features. Because of this, it takes time to estimate the initial position. For the same reason, the proposed algorithm is suitable for UAVs flying long distances outdoors. So, it can be applied to long-distance goods delivery systems that use UAVs, and reconnaissance in the air in areas where attacks such as GPS spoofing can occur. However, due to the characteristics of the features used, the current algorithm has limitations in indoor navigation. If the proposed algorithm is improved by learning various classes that are likely to be indoors, it can be extended to indoor navigation in the future. For the future work, various classes such as roads which are mostly present in various environments will be used as feature information as well. Learning of various classes may improve the accuracy and the convergence speed of the proposed algorithm. Additionally, we expect that the improved algorithm will operate robustly in various environments. In such settings, the algorithm can be verified to be more robust with additional experiments in complex urban environments.





\bibliographystyle{IEEEtran}
\bibliography{references}

\end{document}